\documentclass[runningheads]{llncs}

\usepackage{eccv}

\usepackage{eccvabbrv}

\usepackage{graphicx}
\usepackage{booktabs}
\usepackage{tabularx}
\usepackage{makecell}

\usepackage[accsupp]{axessibility}  
\usepackage{algorithm}
\usepackage{algorithmic}
\usepackage{multirow}
\usepackage{subcaption}
\usepackage{svg}
\usepackage{pdfpages}
\usepackage{amsmath}

\usepackage{hyperref}

\usepackage{orcidlink}

\begin{document}

\title{FMCL: Class-Aware Client Clustering with Foundation Model Representations for Heterogeneous Federated Learning} 

\titlerunning{FMCL}

\author{Mahad Ali\inst{1} \and
Laura J. Brattain\inst{1}}

\authorrunning{M.~Ali et al.}

\institute{University of Central Florida, Orlando FL 32816, USA
\email{\{mahad.ali,laura.brattain\}@ucf.edu}}

\maketitle

\begin{abstract}
  Federated Learning (FL) enables collaborative model training across distributed clients without sharing raw data, yet its performance deteriorates under statistical heterogeneity. Clustered Federated Learning addresses this challenge by grouping similar clients and training separate models per cluster. However, existing clustering strategies often rely on raw data statistics, model parameters, or heuristic similarity measures that fail to capture class-level semantic structure across heterogeneous domains and frequently require iterative coordination. We propose FMCL, a one-shot, class-aware client clustering framework that leverages foundation model representations to construct semantic client signatures. Using a frozen foundation model, FMCL computes class-level embedding prototypes for each client and measures similarity via cosine distance between their class-aware representations. Clustering is performed once prior to training, introducing no additional communication during federated optimization and remaining agnostic to the downstream model architecture. Extensive experiments across heterogeneous benchmarks demonstrate that FMCL improves federated performance and yields more stable clustering behavior compared to existing clustering-based methods under non-identically distributed data partitioning.
\end{abstract}

\section{Introduction}
\label{sec:intro}

Most modern machine learning models benefit from training on large-scale and diverse datasets.
However, in many practical scenarios, particularly in healthcare and mobile applications, privacy regulations and data governance constraints prevent centralizing data for training.

Federated Learning (FL) addresses this challenge by training models collaboratively across distributed clients while keeping data local \cite{fedavg}. In FL, clients perform local optimization on their private datasets and share model updates with a central server for aggregation. This paradigm has enabled privacy-preserving deployment of machine learning systems, including applications such as personalized keyboard prediction on mobile devices.

Despite its advantages, FL faces a fundamental challenge: non-independent and non-identically distributed (non-IID) data. In real-world deployments, client data distributions often differ significantly due to demographic, device, or site variations. Such heterogeneity can severely degrade the performance of classical global aggregation methods such as FedAvg. When client distributions diverge substantially, training a single global model often leads to suboptimal solutions that perform poorly for many clients.

Clustered Federated Learning (CFL) has emerged as a principled solution to this problem. Instead of assuming all clients share one common distribution, CFL partitions clients into groups with similar data distributions and trains a separate model per cluster.

Clustered FL therefore provides a middle ground between:
\begin{itemize}
    \item Global training (single shared model), and
    \item Local training (fully personalized models).
\end{itemize}

This is particularly important in settings where the overall population consists of multiple latent sub-distributions such as different disease subtypes, imaging devices, or acquisition protocols. 
In such cases, enforcing a single global model can fundamentally limit performance, whereas clustering allows semantically similar clients to share statistical strength while remaining distinct from dissimilar groups.

Early CFL methods cluster clients using model gradients computed after several rounds of global training \cite{cfl}. However, gradient-based clustering suffers from two major limitations:
\begin{itemize}
    \item Instability: Gradients early in training can be noisy and unstable, leading to unreliable clustering.
    \item Hyperparameter sensitivity: Clustering decisions often depend on manually tuned thresholds, with limited guidance on robust selection.
\end{itemize}

Other approaches, such as IFCA \cite{ifca}, assume the number of clusters is known a priori and require iterative reassignment of clients, increasing computational complexity.

More recent one-shot methods attempt to cluster clients using data-derived signatures instead of gradients. PACFL uses truncated SVD on raw client data \cite{pacfl}, while FECFL clusters clients using mean feature embeddings \cite{10889437}. These approaches represent important advances by avoiding iterative gradient-based clustering.

However, important limitations remain:
\begin{itemize}
    \item PACFL relies on raw pixel statistics, which may not capture task-relevant semantic structure and may generalize poorly across domains.
    \item FECFL uses only mean feature embeddings, ignoring distributional structure within each class and potentially failing when client differences are subtle.
    \item Many methods continue to depend on heuristic distance thresholds without principled mechanisms for determining the number of clusters.
\end{itemize}
These limitations motivate a more robust and semantically meaningful approach to client representation and clustering.

\begin{figure}
    \centering
    \includegraphics[width=1.0\linewidth]{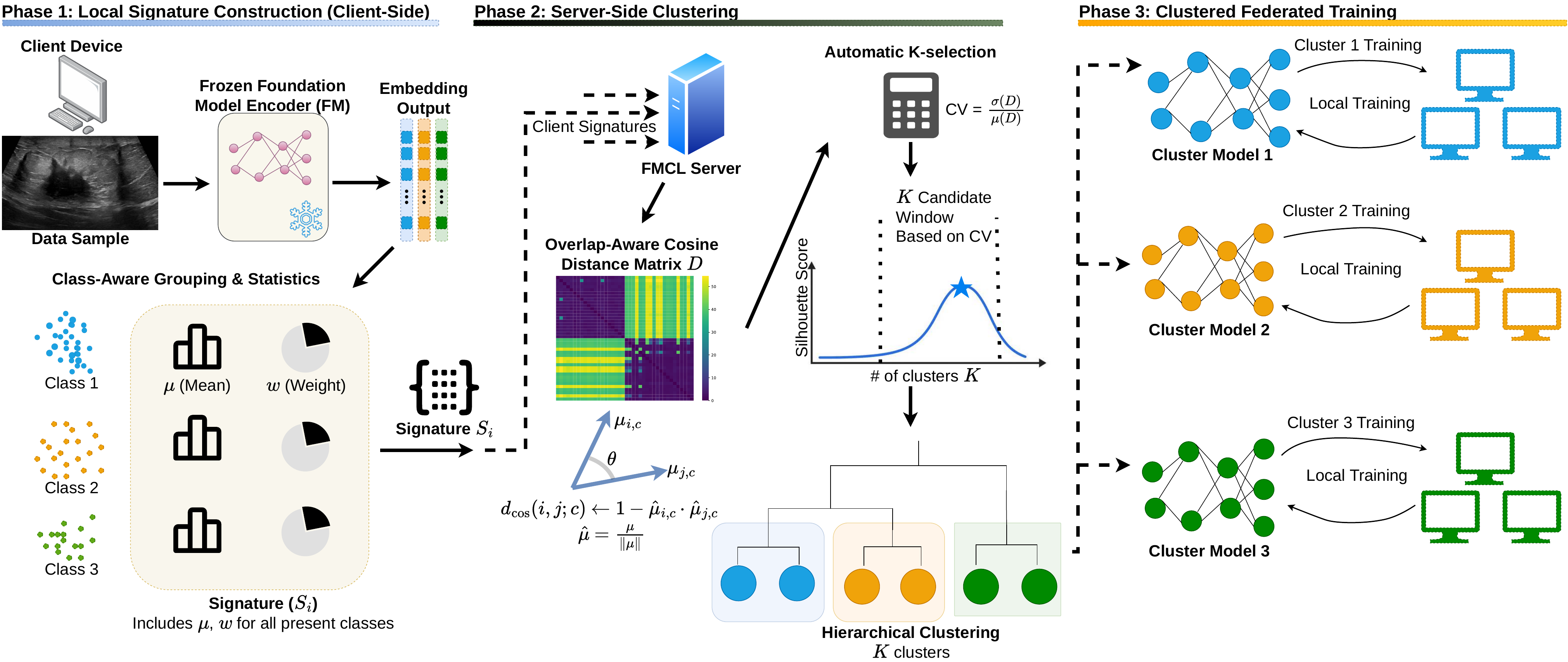}
    \caption{\textbf{The FMCL Framework.} (Left) Clients extract class-aware semantic signatures using a frozen domain-specific foundation model. (Center) The server computes an overlap-aware cosine distance matrix and selects the number of clusters \(K\) via CV-guided silhouette analysis. (Right) One-shot clustering is followed by independent federated training within each cluster.}
    \label{fig:fmcl_overview}
\end{figure}

In this work, we propose \textbf{Foundation Model-based Client Clustering (FMCL)}, a one-shot clustering framework that leverages pretrained foundation model (FM) embeddings to construct class-aware representations of client data distributions.
An overview of the FMCL pipeline is shown in fig \ref{fig:fmcl_overview}.

Unlike prior methods, FMCL:
\begin{itemize}
    \item Constructs class-conditional semantic signatures using mean embeddings extracted from a frozen foundation model,
    \item Accounts for class overlap between clients to avoid misleading similarity when shared classes are sparse,
    \item Computes overlap-aware cosine distances between client signatures,
    \item Applies hierarchical clustering to form client groups,
    \item Provides an automatic mechanism for selecting the number of clusters.
\end{itemize}

By leveraging semantically rich FM embeddings and explicitly modeling distributional structure, FMCL yields more stable and meaningful client clusters across heterogeneous datasets.

We evaluate FMCL on multiple image classification benchmarks, including two medical imaging datasets and one natural image dataset. Across diverse heterogeneity settings, FMCL consistently outperforms global aggregation and prior clustered FL methods.

\section{Related Work}
\label{sec:related_works}

\paragraph{Federated Learning under Statistical Heterogeneity.}
Federated Learning (FL) enables collaborative model training without centralizing data \cite{fedavg}. However, numerous studies have shown that classical global aggregation methods such as FedAvg degrade significantly under statistical heterogeneity (non-IID data) \cite{DBLP:journals/corr/abs-1806-00582, hsu2019measuringeffectsnonidenticaldata}. 

To address this, several methods attempt to reduce client drift during training. FedProx introduces a proximal regularization term \cite{fedprox}; SCAFFOLD leverages control variates to correct update bias \cite{scaffold}; FedOpt incorporates adaptive server-side optimizers \cite{fedopt}; and FedNova normalizes local updates to mitigate objective inconsistency \cite{fednova}. While these approaches improve convergence stability, they still rely on learning a single global model and may struggle when client data distributions form distinct subpopulations.

\paragraph{Personalized Federated Learning.}
Another direction focuses on personalization. Methods such as Per-FedAvg \cite{perfedavg} and local fine-tuning strategies \cite{yu2022salvagingfederatedlearninglocal} aim to adapt global models to individual clients. Although effective in some settings, these approaches assume independence across clients and fail to exploit potential shared structure among subsets of clients.

\paragraph{Clustered Federated Learning.}
Clustered Federated Learning (CFL) explicitly models client subpopulations by grouping clients with similar data distributions. The original CFL method clusters clients using gradient similarity after several rounds of FedAvg \cite{cfl}. However, this approach depends on gradient stability and threshold selection, making clustering sensitive to optimization dynamics.

IFCA assigns clients to a predefined number of clusters via iterative reassignment based on local loss \cite{ifca}. While effective under correct assumptions, it requires prior knowledge of the number of clusters and incurs increased computational cost.

More recent one-shot methods attempt to cluster clients directly using data-derived representations. PACFL constructs client signatures using truncated SVD of raw images and applies hierarchical clustering based on principal angles \cite{pacfl}. 
FECFL instead proposes clustering clients using feature representations extracted from a lightweight encoder and reports that effective clustering can be achieved even with randomly initialized models, reducing the need for pretrained representations \cite{10889437}.
These approaches remove the need for iterative gradient-based clustering.

However, key limitations remain. PACFL relies on raw pixel statistics, which may not capture semantically meaningful structure and can be highly sensitive to threshold selection. FECFL uses only mean feature embeddings, potentially overlooking distributional variability within classes and subtle inter-client differences. Furthermore, existing one-shot methods still depend on heuristic thresholds or manual selection of the number of clusters.

\paragraph{Foundation Model Representations.}
Foundation models (FMs) trained on large-scale datasets provide semantically rich and transferable representations across tasks and domains. Leveraging these representations for client characterization offers a promising direction for robust clustering under heterogeneity.

In contrast to prior clustered FL approaches, our proposed method constructs class-aware client signatures using FM embeddings, explicitly models distributional structure via cosine distances, incorporates class overlap awareness, and introduces a principled strategy for automatic cluster selection.
\section{Methodology}
\label{sec:methodology}

In this section, we will start with preliminaries about FL, provide an in-depth explanation about our proposed algorithm, followed by details about the experiments that we performed.

\subsection{Preliminaries}
The baseline FL algorithm, FedAvg \cite{fedavg}, aims to find global model parameters \(\theta\) that minimize the weighted average loss \(\sum_{i=1}^N p_i f_\theta(d_i)\) over clients' local datasets \(d_i\), where \(1 \leq i \leq N\). 
Here, \(p_i\) is the relative weight of client \(i\), typically set as the data proportion, i.e., \(p_i = \frac{|d_i|}{\sum_{j=1}^N |d_j|}\).

FedAvg starts with a global model, \(\theta_0\), and updates it over multiple rounds.
At the start of each round \(r\), where \(1 \leq r \leq R\), the server sends the global model \(\theta_{r - 1}\) to the clients.
The clients update the received model by running Stochastic Gradient Descent (SGD) on their local data.
Each client sends its updated local model \(\theta_i^r\) to the server, which performs weighted averaging of the client models to obtain the new global model \(\theta_r\).
Algorithm \ref{algo:fedavg} provides more details about the baseline federated learning algorithm.

\begin{algorithm}
    \caption{Baseline Federated Learning algorithm. \(N\): number of clients, \(s\): client sampling fraction, \(R\): number of rounds, \(E\): local epochs, \(B\): batch size, \(d_i\): local data of client \(i\), \(\eta\): learning rate.}
    \label{algo:fedavg}
    \begin{algorithmic}
        \STATE \textbf{Initialize:} Global model parameters \(\theta_0\)

        \FOR{$r = 1$ to $R$}
            \STATE $m \leftarrow \max(s \cdot N, 1)$
            \STATE $S_r \leftarrow$ randomly sample $m$ clients
            \STATE Send global model \(\theta_{r-1}\) to all clients in \(S_r\)
            
            \FOR{each client $i \in S_r$ \textbf{(in parallel)}}
                \STATE $\theta_{i}^{r} \leftarrow \theta_{r-1}$
                \STATE Split \(d_i\) into batches of size \(B\)

                \FOR{$e = 1$ to $E$}
                    \FOR{each batch $b$ in \(d_i\)}
                        \STATE $\theta_{i}^{r} \leftarrow \theta_{i}^{r} - \eta \nabla_{\theta} f_{\theta_{i}^{r}}(b)$
                    \ENDFOR
                \ENDFOR

                \STATE Send \(\theta_i^r\) to server
            \ENDFOR

            \STATE \textbf{Server aggregates:}
            \STATE $\theta_r \leftarrow \frac{1}{\sum_{i \in S_r} |d_i|} \sum_{i \in S_r} |d_i| \cdot \theta_i^r$
        \ENDFOR
    \end{algorithmic}
\end{algorithm}

While FedAvg is the baseline Federated Learning algorithm, other single global model aggregation approaches have proposed solutions to the problems associated with FedAvg. 
For example, FedProx mitigates heterogeneity-related challenges by introducing a regularization term in the clients’ loss function, which penalizes divergence from the global weights.

While global model aggregation strategies have shown promising results, as demonstrated in the work CFL \cite{cfl}, they struggle in situations where client datasets might belong to \(K \leq N\) different distributions instead of following one global distribution.
CFL's proposed approach groups clients with a similar data distribution together into clusters, wherein each cluster runs its model aggregation separately from other clusters.
If \(K = N\), then all clients train the model locally on their datasets. If \(K=1\), then clients train the model using a global aggregation strategy (e.g. FedAvg, FedProx).
\subsection{FMCL}
In this subsection, we go over the details of our proposed federated clustering approach.
We start with client signature construction, which are then used to compute pairwise distances between clients. These pairwise distances are then used to obtain clusters using hierarchical clustering.

\subsubsection{Client Signature Construction}
Let $\phi(\cdot)$ denote the frozen pretrained FM encoder.
For client $i$ with dataset $d_i$, we extract feature embeddings 
$z_{i,n} = \phi(x_{i,n})$ for all samples $x_{i,n} \in d_i$.
For each class $c$ present in $d_i$, we compute:

\begin{align}
\mu_{i,c} &= \frac{1}{|d_{i,c}|} \sum_{x \in d_{i,c}} \phi(x) \\
w_{i,c} &= \frac{|d_{i,c}|}{|d_i|}
\end{align}

The client signature is defined as:
\[
\mathcal{S}_i = \{ (\mu_{i,c}, w_{i,c}) \}_{c \in \mathcal{C}_i}
\]
where $\mathcal{C}_i$ denotes the set of classes present on client $i$.

Algorithm \ref{algo:signature} provides more details regarding client signature construction.

\begin{algorithm}
\caption{Client Signature Construction (Class-Aware, Element-Wise Statistics)}
\label{algo:signature}
\begin{algorithmic}
\STATE \textbf{Input:} Client dataset $d_i$, pretrained encoder $\phi$, embedding dimension $d$
\STATE \textbf{Output:} Client signature $\mathcal{S}_i = \{(\mu_{i,c}, w_{i,c})\}_{c \in \mathcal{C}_i}$

\STATE $\mathcal{S}_i \leftarrow \emptyset$
\STATE $\mathcal{C}_i \leftarrow$ set of classes present in $d_i$

\FOR{each class $c \in \mathcal{C}_i$}
    \STATE $d_{i,c} \leftarrow \{(x,y)\in d_i : y=c\}$ \;\;\; and \;\;\; $n_{i,c} \leftarrow |d_{i,c}|$
    \STATE Extract embeddings $Z_{i,c} \leftarrow \{\phi(x) \in \mathbb{R}^d : (x,y)\in d_{i,c}\}$

    \STATE \textbf{Element-wise mean:}
    \[
    \mu_{i,c} \leftarrow \frac{1}{n_{i,c}} \sum_{z \in Z_{i,c}} z
    \]

    \STATE \textbf{Class weight:} \;\; $w_{i,c} \leftarrow \frac{n_{i,c}}{|d_i|}$

    \STATE $\mathcal{S}_i \leftarrow \mathcal{S}_i \cup \{(\mu_{i,c}, w_{i,c})\}$
\ENDFOR

\STATE \textbf{return} $\mathcal{S}_i$
\end{algorithmic}
\end{algorithm}

\subsubsection{Pairwise Distance Matrix Construction}

We utilize the client signatures $\mathcal{S}_i$ obtained from Algorithm~\ref{algo:signature} to construct a pairwise distance matrix used for clustering.

For a pair of clients $(i,j)$, let $\mathcal{C}_i$ and $\mathcal{C}_j$ denote the sets of classes present in their respective local datasets. 
We define $\mathcal{C}_{ij} = \mathcal{C}_i \cap \mathcal{C}_j$ as the set of overlapping classes.

If $\mathcal{C}_{ij} \neq \emptyset$, we compute a weighted cosine distance between their class-conditional embedding distributions.
The contribution of each overlapping class $c$ is weighted by $\min(w_{i,c}, w_{j,c})$, where $w_{i,c}$ represents the class proportion of class $c$ on client $i$. 
The resulting distance is further scaled inversely by the total overlap 
$\Omega_{ij} = \sum_{c \in \mathcal{C}_{ij}} \min(w_{i,c}, w_{j,c})$ 
to penalize pairs with limited shared label support.
Without overlap-aware scaling, two clients that share only a small subset of classes could appear close due to agreement on limited label support, despite having substantially different overall distributions.

If $\mathcal{C}_{ij} = \emptyset$, the class-conditional metric is undefined. 
In this case, we assign a large distance $D_{\text{big}}$ derived from a high percentile of the observed finite pairwise distances. 
This ensures that non-overlapping clients are treated as dissimilar while avoiding the use of a dataset-dependent absolute constant.

Algorithm~\ref{algo:distmat} summarizes the full distance matrix construction.
 
\begin{algorithm}
\caption{Pairwise Distance Matrix Construction (Class- and Overlap-Aware)}
\label{algo:distmat}
\begin{algorithmic}
\STATE \textbf{Input:} Client signatures $\{\mathcal{S}_i\}_{i=1}^N$, hyperparameters $\alpha, \beta$, small constant $\varepsilon$
\STATE \textbf{Output:} Distance matrix $D \in \mathbb{R}^{N \times N}$

\STATE Initialize $D \leftarrow \infty \cdot \mathbf{1}_{N \times N}$
\STATE $\mathcal{C}_i \leftarrow$ set of classes present in $\mathcal{S}_i$

\FOR{each pair $(i,j)$ with $i < j$}
    \STATE $\mathcal{C}_{ij} \leftarrow \mathcal{C}_i \cap \mathcal{C}_j$
    \IF{$\mathcal{C}_{ij} = \emptyset$}
        \STATE $D_{ij} \leftarrow \infty$
         \STATE \textbf{continue}
    \ENDIF
    
    \STATE $\Omega_{ij} \leftarrow 0$, \;\; $num \leftarrow 0$, \;\; $den \leftarrow 0$
    \FOR{each class $c \in \mathcal{C}_{ij}$}
        \STATE $\tilde w_{ij,c} \leftarrow \min(w_{i,c}, w_{j,c})$
        \STATE $\Omega_{ij} \leftarrow \Omega_{ij} + \tilde w_{ij,c}$
        
        \STATE $d_{\cos}(i,j;c) \leftarrow 1 - \frac{{{\mu_{i,c}^\top}} \cdot \mu_{j,c}}{\lVert\mu_{i,c}\rVert \lVert\mu_{j,c}\rVert + \varepsilon}$
        
        \STATE $num \leftarrow num + \tilde w_{ij,c} \cdot d_{\cos}(i,j;c)$
        \STATE $den \leftarrow den + \tilde w_{ij,c}$
    \ENDFOR
    
    \STATE $d_{\text{cap}}(i,j) \leftarrow \frac{num}{den + \varepsilon}$
    \STATE $m_{ij} \leftarrow \min\left((\max(\Omega_{ij}, \varepsilon))^{-\alpha}, \beta\right)$
    \STATE $D_{ij} \leftarrow d_{\text{cap}}(i,j) \cdot m_{ij}$
    \STATE $D_{ji} \leftarrow D_{ij}$
\ENDFOR

\STATE Let $\mathcal{V} \leftarrow \{D_{ij} : i\neq j \text{ and } D_{ij} < \infty\}$
\STATE Compute $P_{95}, P_{99}$ as the 95th and 99th percentiles of $\mathcal{V}$
\STATE $D_{\text{big}} \leftarrow \min(2\cdot P_{95}, P_{99})$
\FOR{each pair $(i,j)$ with $i\neq j$}
    \IF{$D_{ij} = \infty$}
        \STATE $D_{ij} \leftarrow D_{\text{big}}$
    \ENDIF
\ENDFOR

\STATE Set $\mathrm{diag}(D) \leftarrow 0$
\STATE \textbf{return} $D$
\end{algorithmic}
\end{algorithm}

\paragraph{Hyperparameter settings.}
Unless otherwise stated, we use $\alpha=1.0$, $\beta=100$, and $\varepsilon=10^{-3}$ for overlap-aware scaling.

\subsubsection{Client clustering from the distance matrix}
Given the pairwise distance matrix $D \in \mathbb{R}^{N \times N}$ (Algorithm~\ref{algo:distmat}), we form client clusters using agglomerative hierarchical clustering.
We initialize each client as its own cluster and iteratively merge the closest pair of clusters according to a chosen linkage criterion.
Clustering stops either when a distance threshold is exceeded (threshold-based clustering) or when a target number of clusters $K$ is reached (fixed-$K$ clustering).
Algorithm~\ref{algo:clustering} summarizes this procedure.

\begin{algorithm}[t]
\caption{Agglomerative clustering from pairwise distances}
\label{algo:clustering}
\begin{algorithmic}[1]
\STATE \textbf{Input:} Distance matrix $D \in \mathbb{R}^{N \times N}$, linkage $\ell \in \{\texttt{single},\texttt{complete},\texttt{average}\}$,
\STATE \hspace{1.35cm} \textbf{either} threshold $\Theta$ \textbf{or} target cluster count $K$
\STATE \textbf{Output:} Client clusters $\mathcal{C} = \{C_1,\dots,C_{|\mathcal{C}|}\}$

\STATE Initialize clusters: $\mathcal{C} \leftarrow \big\{\{1\}, \{2\}, \dots, \{N\}\big\}$
\STATE Define cluster-to-cluster distance $\Delta_\ell(C_a, C_b)$ using $D$ and linkage $\ell$
\WHILE{True}
    \STATE $(C_p, C_q) \leftarrow \arg\min\limits_{C_a \neq C_b \in \mathcal{C}} \Delta_\ell(C_a, C_b)$
    \STATE $\delta^\star \leftarrow \Delta_\ell(C_p, C_q)$
    
    \IF{$\Theta$ is provided \textbf{and} $\delta^\star > \Theta$}
        \STATE \textbf{break} \COMMENT{threshold-based stopping}
    \ENDIF
    
    \IF{$K$ is provided \textbf{and} $|\mathcal{C}| = K$}
        \STATE \textbf{break} \COMMENT{fixed-$K$ stopping}
    \ENDIF
    
    \STATE Merge: $C_{\text{new}} \leftarrow C_p \cup C_q$
    \STATE Update: $\mathcal{C} \leftarrow \big(\mathcal{C} \setminus \{C_p, C_q\}\big) \cup \{C_{\text{new}}\}$
\ENDWHILE
\STATE \textbf{return} $\mathcal{C}$
\end{algorithmic}
\end{algorithm}

\paragraph{Clustering configuration.}
We use average linkage in all reported experiments.

\subsubsection{Automatic cluster selection}

The number of underlying client distributions is generally unknown in practical federated settings. 
We therefore adopt a data-driven yet deterministic procedure to select the number of clusters.

We first estimate the overall heterogeneity level from the dispersion of the pairwise distance matrix.
Specifically, we compute the coefficient of variation (CV) of the off-diagonal distances:
\[
\text{CV} = \frac{\mathrm{std}(D_{ij})}{\mathrm{mean}(D_{ij})}, \quad i \neq j.
\]
Higher dispersion indicates greater variability across client distributions.

Using fixed, dataset-independent thresholds on CV, we define an adaptive candidate window for the number of clusters and evaluate clustering quality via the silhouette score.
The selected cluster count corresponds to the dominant local maximum of the silhouette curve.
Algorithm~\ref{algo:autok} summarizes the procedure.

\begin{algorithm}[t]
\caption{Automatic Selection of Cluster Count}
\label{algo:autok}
\begin{algorithmic}[1]
\STATE \textbf{Input:} Distance matrix $D$, linkage $\ell$, maximum cluster count $K_{\max}$
\STATE \textbf{Output:} Selected cluster count $K^\star$

\STATE Compute coefficient of variation $CV$ of the off-diagonal entries of $D$

\STATE Determine candidate window $\mathcal{K}_{\text{window}}$ using fixed CV thresholds

\FOR{each $K \in \mathcal{K}_{\text{window}}$}
    \STATE Obtain clusters $\mathcal{C}_K$ using Algorithm~\ref{algo:clustering}
    \STATE Compute silhouette score $S(K)$ using $D$
\ENDFOR

\STATE Identify local maxima of $S(K)$ within $\mathcal{K}_{\text{window}}$

\IF{local maxima exist}
    \STATE $K^\star \leftarrow$ largest local maximum
\ELSE
    \STATE $K^\star \leftarrow \arg\max_{K \in \{1,\dots,K_{\max}\}} S(K)$
\ENDIF

\STATE \textbf{return} $K^\star$
\end{algorithmic}
\end{algorithm}

\paragraph{Silhouette-based cluster validation.}
We use the silhouette score as an internal cluster validity metric because it directly measures the separation and compactness of clusters under a precomputed distance matrix.
For each client $i$, the silhouette value is defined as
\[
s(i) = \frac{b(i) - a(i)}{\max\{a(i), b(i)\}},
\]
where $a(i)$ is the average distance between client $i$ and other clients in the same cluster, and $b(i)$ is the minimum average distance between client $i$ and clients in a different cluster.
The overall silhouette score is the mean of $s(i)$ across all clients.

Since our method constructs a distance matrix that reflects class-conditional distribution differences, silhouette evaluation is well-suited for assessing the resulting cluster structure without requiring access to ground-truth client labels.
\section{Experimental Setup}
\label{sec:experimental_setup}
This section describes the experimental protocol used to evaluate our method, including datasets, data partitioning strategy, baselines, model architectures, implementation details, and evaluation metrics.

\subsection{Datasets}
We evaluate our approach on three image classification datasets:

\begin{itemize}
    \item \textbf{Breast Ultrasound Image (BUSI):} A dataset containing 780 breast ultrasound images classified into three categories: benign, malignant, and normal.
    \item \textbf{LungHist700:} A histopathology dataset consisting of 700 lung tissue images categorized into squamous cell carcinoma, adenocarcinoma, and normal.
    \item \textbf{Imagenette:} A 10-class subset of ImageNet commonly used for benchmarking image classification models.
\end{itemize}

These datasets cover both medical imaging (BUSI, LungHist700) and natural image classification (Imagenette), enabling evaluation across domains with varying levels of visual complexity and semantic structure.

\subsection{Data Partitioning}

To simulate statistical heterogeneity across clients, we partition each dataset using a Dirichlet distribution over class labels.
The Dirichlet concentration parameter $\alpha$ controls the degree of heterogeneity (smaller values induce stronger label skew); unless stated otherwise, we use $\alpha = 0.1$.

We distribute LungHist700 and BUSI across 20 clients, and Imagenette across 30 clients.
Fewer clients are used for the smaller datasets (LungHist700 and BUSI) to ensure sufficient data per client.

To account for randomness in data partitioning and model initialization, all experiments are repeated with fixed random seeds.
Results for LungHist700 and BUSI are averaged over 5 seeds, and results for Imagenette are averaged over 3 seeds.

\subsection{Baselines}

We compare FMCL against five baselines commonly used in heterogeneous federated learning: FedAvg \cite{fedavg}, the standard global aggregation algorithm; FedProx \cite{fedprox}, which introduces a proximal regularization term to mitigate client drift; CFL \cite{cfl}, which clusters clients based on gradient similarity during training; and FECFL \cite{10889437} , which clusters clients using feature embeddings extracted from an encoder.

To ensure a fair comparison, we evaluate clustering-based baselines under comparable representation settings. 
Specifically, both FMCL and FECFL are evaluated using the same frozen foundation model embeddings. 
Additionally, following the original FECFL design, we also evaluate FECFL using randomly initialized encoders to assess the impact of representation quality on clustering performance.

PACFL was originally proposed using raw image inputs. 
For completeness, we evaluate PACFL both in its original form and with foundation model embeddings used as input to its SVD-based subspace signature construction.

\subsection{Model Architectures}

For federated training, we use ResNet9 for the smaller medical imaging datasets (LungHist700 and BUSI) and ResNet18 for Imagenette.
ResNet9 provides a lightweight architecture suitable for smaller datasets with only three classes, while the larger and more diverse Imagenette dataset benefits from the increased capacity of ResNet18. 
All models are trained from scratch with random initialization.

For client signature construction, we use domain-specific foundation models:
CHIEF \cite{chief}, a pathology foundation model trained on large-scale histopathology images;
USFM \cite{USFM}, an ultrasound foundation model for medical ultrasound analysis;
and ViT-Tiny \cite{vit}, a lightweight vision transformer pretrained on ImageNet.
The same feature extractors are used for FECFL and PACFL (when evaluated with FM embeddings) to ensure a controlled comparison.

\subsection{Implementation Details}

All federated learning methods are trained for 100 communication rounds with one local epoch per round.
A full participation rate (100\%) is used for all experiments.

We use stochastic gradient descent (SGD) as the optimizer.
The learning rate is set to 0.0001 for LungHist700 and BUSI, and 0.01 for Imagenette.

For automatic cluster selection (Algorithm~\ref{algo:autok}), we compute the coefficient of variation (CV) of the off-diagonal entries of the distance matrix $D$.
If $\mathrm{CV} < 0.35$, we evaluate $K \in \{1,2,3\}$;
if $0.35 \le \mathrm{CV} < 0.70$, we evaluate $K \in \{2,\dots,6\}$;
otherwise, we evaluate $K \in \{3,\dots,10\}$.
These thresholds are fixed across all datasets.

\subsection{Evaluation Metrics}

We evaluate federated learning performance using classification accuracy, macro-averaged F1-score, and Area Under the Receiver Operating Characteristic Curve (AUC-ROC).

For each method, experiments are repeated across multiple random seeds.
At each communication round, we compute the evaluation metrics and then average them across seeds.
Performance curves therefore represent the mean trajectory over multiple runs.

For quantitative comparison, we report the mean $\pm$ standard deviation of the best average performance achieved over communication rounds.
Specifically, for each method we first average performance across seeds at every round, and then select the round corresponding to the highest mean validation metric.
\section{Results}
\label{sec:results}

As outlined in Section~\ref{sec:experimental_setup}, we evaluate our method on three datasets: LungHist700, BUSI, and Imagenette. 
To ensure a fair comparison among one-shot clustering approaches (FMCL, FECFL, and PACFL), we fix the number of clusters \(K\) for each dataset. 
Guided by our automated cluster selection algorithm, we set \(K=3\) for LungHist700 and BUSI, and \(K=5\) for Imagenette. 
Given the total number of clients in each dataset (\(N=20\) for BUSI and LungHist700, \(N=30\) for Imagenette), these choices ensure sufficiently populated clusters while maintaining meaningful inter-cluster separation.

\begin{figure}
    \centering
    \includegraphics[width=1.0\linewidth]{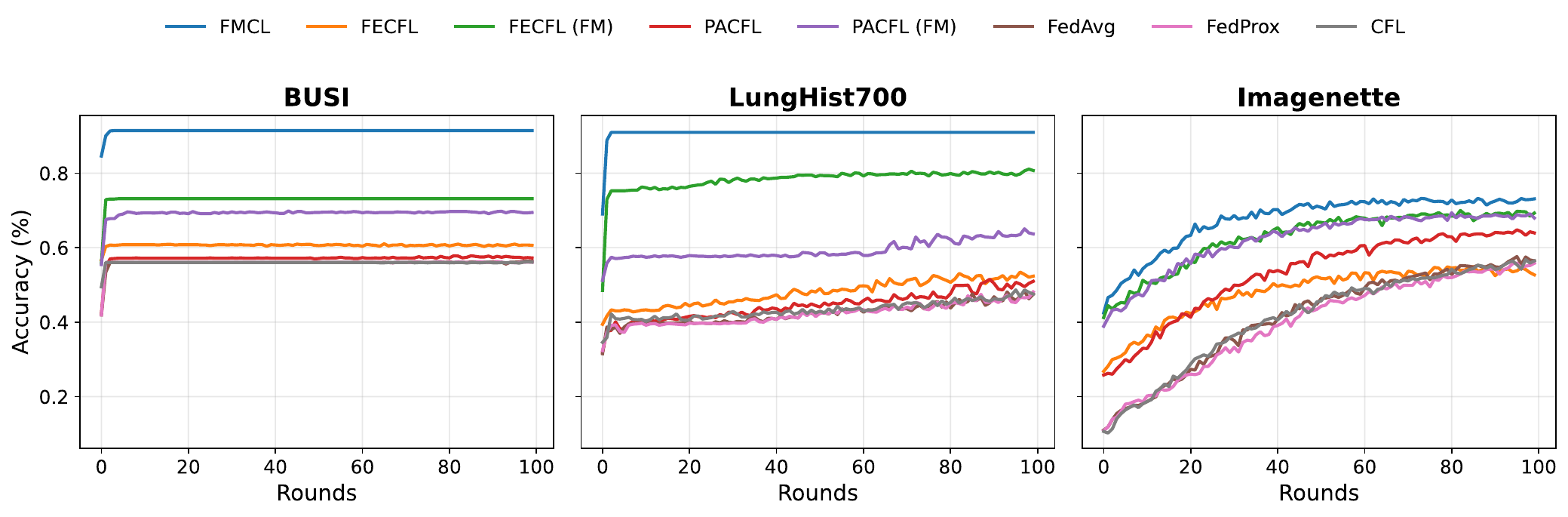}
    \caption{Performance comparison of federated learning strategies across three datasets. FMCL achieves higher accuracy and faster convergence than competing methods.}
    \label{fig:accuracy_v_rounds}
\end{figure}

Figure~\ref{fig:accuracy_v_rounds} reports the mean test accuracy across FL communication rounds. 
Across all datasets, FMCL consistently outperforms both global aggregation methods and prior clustering-based approaches.

We also observe that incorporating foundation model (FM) embeddings significantly improves clustering-based methods. 
Both PACFL and FECFL benefit from replacing raw image features with FM embeddings, highlighting the importance of high-quality semantic representations for capturing client-level distributional heterogeneity.

Table~\ref{tab:main_results} summarizes the best average performance (mean $\pm$ std) across runs, selected based on validation loss. 
We additionally report an ablation of FMCL without the proposed overlap-aware scaling in the distance computation. 
Removing the overlap term consistently degrades performance across datasets, highlighting the importance of accounting for shared class support when measuring client similarity.

We also report results using our automatic cluster selection algorithm (Algorithm~\ref{algo:autok}). 
Across all datasets, Auto-$K$ further improves performance and reduces variance compared to fixed-\(K\) clustering, demonstrating the robustness of our cluster selection strategy.

Despite the strong improvements obtained by integrating FM embeddings into existing clustering methods, FMCL consistently achieves the best overall performance. 
In particular, FMCL (Auto-$K$) obtains the highest accuracy, F1, and AUC on all three datasets, substantially outperforming both global aggregation baselines and prior clustering-based approaches.

\begin{table}[t]
\centering
\caption{Comparison of federated learning strategies across datasets (mean $\pm$ std). Best results are bold. FMCL consistently achieves the strongest performance across all datasets, with Auto-$K$ providing further improvements.}
\label{tab:main_results}
\scriptsize
\setlength{\tabcolsep}{3pt}
\renewcommand{\arraystretch}{1.05}

\begin{subtable}{\linewidth}
\centering
\caption*{\textbf{BUSI}}
\begin{tabular}{lccc}
\toprule
\textbf{Strategy} & \textbf{Acc} & \textbf{F1} & \textbf{AUC} \\
\midrule
FedAvg      & 56.34$\pm$0.90 & 24.82$\pm$1.38 & 59.58$\pm$2.79 \\
FedProx     & 56.18$\pm$0.71 & 24.39$\pm$0.68 & 58.55$\pm$2.19 \\
CFL         & 56.18$\pm$0.71 & 24.38$\pm$0.96 & 61.06$\pm$3.20 \\
\midrule
PACFL       & 57.25$\pm$0.67 & 27.87$\pm$2.03 & 61.89$\pm$3.86 \\
PACFL (FM) & 69.86$\pm$8.80 & 51.69$\pm$13.45 & 72.87$\pm$10.27 \\
\midrule
FECFL       & 61.04$\pm$4.21 & 44.08$\pm$9.22 & 72.11$\pm$6.52 \\
FECFL (FM)       & 73.15$\pm$6.07 & 56.27$\pm$8.89 & 75.09$\pm$4.16 \\
\midrule
FMCL (no overlap) & 90.12$\pm$5.21 & 84.34$\pm$13.30 & 94.44$\pm$1.62 \\
FMCL     & 91.43$\pm$2.59 & 88.41$\pm$5.52 & 93.93$\pm$1.36 \\
FMCL (Auto-$K$) & \textbf{91.68$\pm$2.52} & \textbf{88.72$\pm$5.61}  & \textbf{95.87$\pm$2.03} \\
\bottomrule
\end{tabular}
\end{subtable}

\vspace{0.6em}

\begin{subtable}{\linewidth}
\centering
\caption*{\textbf{LungHist700}}
\begin{tabular}{lccc}
\toprule
\textbf{Strategy} & \textbf{Acc} & \textbf{F1} & \textbf{AUC} \\
\midrule
FedAvg      & 48.38$\pm$6.34 & 32.03$\pm$9.21 & 73.88$\pm$4.09 \\
FedProx     & 48.10$\pm$4.67 & 32.81$\pm$7.59 & 74.72$\pm$4.15 \\
CFL         & 48.66$\pm$6.18 & 32.42$\pm$8.73 & 73.96$\pm$4.96 \\
\midrule
PACFL       & 51.51$\pm$4.21 & 38.25$\pm$4.70 & 74.46$\pm$3.60 \\
PACFL (FM) & 65.07$\pm$8.56 & 62.80$\pm$12.67 & 82.34$\pm$6.13 \\
\midrule
FECFL       & 53.45$\pm$3.74 & 42.40$\pm$4.15 & 75.90$\pm$4.72 \\
FECFL (FM)       & 81.09$\pm$8.41 & 81.06$\pm$7.22 & 91.04$\pm$5.38 \\
\midrule
FMCL (no overlap) & 87.65$\pm$5.20 & 84.48$\pm$8.36 & 94.92$\pm$1.51 \\
FMCL     & 90.97$\pm$2.12 & 90.01$\pm$2.75 & 96.27$\pm$1.26 \\
FMCL (Auto-$K$) & \textbf{91.71$\pm$2.66} & \textbf{90.94$\pm$3.30}  & \textbf{96.51$\pm$1.00} \\
\bottomrule
\end{tabular}
\end{subtable}

\vspace{0.6em}

\begin{subtable}{\linewidth}
\centering
\caption*{\textbf{Imagenette}}
\begin{tabular}{lccc}
\toprule
\textbf{Strategy} & \textbf{Acc} & \textbf{F1} & \textbf{AUC} \\
\midrule
FedAvg      & 57.62$\pm$3.13 & 55.61$\pm$3.44 & 91.00$\pm$0.65 \\
FedProx     & 55.84$\pm$3.95 & 53.24$\pm$4.81 & 90.54$\pm$1.07 \\
CFL         & 56.79$\pm$4.07 & 54.49$\pm$5.16 & 90.80$\pm$1.12 \\
\midrule
PACFL       & 64.79$\pm$4.81 & 63.44$\pm$4.95 & 93.00$\pm$1.32 \\
PACFL (FM) &  68.38$\pm$4.19 & 68.00$\pm$4.91 & 94.23$\pm$1.37 \\
\midrule
FECFL       & 54.91$\pm$3.72 & 54.22$\pm$3.37 & 89.22$\pm$1.08 \\
FECFL (FM)       & 69.99$\pm$0.48 & 68.55$\pm$0.15 & 94.37$\pm$0.36 \\
\midrule
FMCL (no overlap) & 71.48$\pm$2.39 & 70.47$\pm$2.15 & 95.26$\pm$0.67 \\
FMCL     & 73.23$\pm$0.98 & 72.93$\pm$1.12 & 95.78$\pm$0.09 \\
FMCL (Auto-$K$) & \textbf{83.10$\pm$1.77} & \textbf{82.82$\pm$1.91} & \textbf{97.58$\pm$0.51} \\
\bottomrule
\end{tabular}
\end{subtable}

\end{table}
\section{Conclusion}

We presented FMCL, a one-shot, class-aware client clustering framework for heterogeneous federated learning. By leveraging semantically rich foundation model representations, modeling class-conditional feature distributions, and incorporating overlap-aware cosine distances, FMCL constructs meaningful client signatures without iterative coordination or additional communication during training. 

Across multiple heterogeneous image benchmarks, FMCL consistently outperformed global aggregation and prior clustered FL methods, while providing a principled mechanism for automatic cluster selection. These results demonstrate that distribution-aware semantic modeling at the client level is a powerful and practical strategy for addressing non-IID federated settings. 

We believe FMCL provides a modular and architecture-agnostic foundation for future research at the intersection of foundation models and federated learning.


%
%
\clearpage

\bibliographystyle{splncs04}
\bibliography{main}
\end{document}